\documentclass[10pt,twocolumn,letterpaper]{article}

\usepackage{wacv}
\usepackage{times}
\usepackage{epsfig}
\usepackage{graphicx}
\usepackage{amsmath}
\usepackage{amssymb}
\usepackage{booktabs}
\usepackage[utf8]{inputenc} % allow utf-8 input
\usepackage[T1]{fontenc}    % use 8-bit T1 fonts
\usepackage{url}            % simple URL typesetting
\usepackage{booktabs}       % professional-quality tables
\usepackage{amsfonts}       % blackboard math symbols
\usepackage{nicefrac}       % compact symbols for 1/2, etc.
\usepackage{microtype}      % microtypography
\usepackage{xcolor}         % colors
\usepackage{algorithm}
\usepackage{algpseudocode}
\usepackage[accsupp]{axessibility}  % Improves PDF readability for those with disabilities.

\usepackage{caption}
\def\wacvPaperID{10} % Enter the WACV Paper ID here

\wacvalgorithmstrack   % Uncomment this line if you are submitting to the Algorithms Track.
\wacvfinalcopy % *** Uncomment this line for the final submission

\ifwacvfinal
\usepackage[breaklinks=true,bookmarks=false]{hyperref}
\else
\usepackage[pagebackref=true,breaklinks=true,colorlinks,bookmarks=false]{hyperref}
\fi

\title{Scalable and Accurate Self-supervised Multimodal Representation Learning without Aligned Video and Text Data}

\author{%
    Vladislav Lialin\textsuperscript{1}\thanks{Work done while at Amazon. Correspondance to Vlad Lialin \texttt{vlialin@cs.uml.edu}}
    \quad
    Stephen Rawls \textsuperscript{2}
    \quad
    David Chan \textsuperscript{2,3}
    \and
    Shalini Ghosh \textsuperscript{2}
    \quad
    Anna Rumshisky \textsuperscript{1,2}
    \quad
    Wael Hamza \textsuperscript{2}
    \and
    \textsuperscript{1} UMass Lowell
    \quad
    \textsuperscript{2} Amazon
    \quad
    \textsuperscript{3} UC Berkeley\\
}

\begin{document}

\maketitle

%%%%%%%%%%%%%%%%%%%%%%%%%%%%%%%%%%%%%%%%%%%%%%%%%%%%%%%%
% Abstract
%%%%%%%%%%%%%%%%%%%%%%%%%%%%%%%%%%%%%%%%%%%%%%%%%%%%%%%%
\begin{abstract}

Scaling up weakly-supervised datasets has shown to be highly effective in the image-text domain and has contributed to most of the recent state-of-the-art computer vision and multimodal neural networks.
However, existing large-scale video-text datasets and mining techniques suffer from several limitations, such as the scarcity of aligned data, the lack of diversity in the data, and the difficulty of collecting aligned data.
Currently popular video-text data mining approach via automatic speech recognition (ASR) used in HowTo100M provides low-quality captions that often do not refer to the video content. Other mining approaches do not provide proper language descriptions (video tags) and are biased toward short clips (alt text).

In this work, we show how recent advances in image captioning allow us to pre-train high-quality video models without any parallel video-text data.
We pre-train several video captioning models that are based on an OPT language model and a TimeSformer visual backbone. We fine-tune these networks on several video captioning datasets.
First, we demonstrate that image captioning pseudolabels work better for pre-training than the existing HowTo100M ASR captions.
Second, we show that pre-training on both images and videos produces a significantly better network (+4 CIDER on MSR-VTT) than pre-training on a single modality.
Our methods are complementary to the existing pre-training or data mining approaches and can be used in a variety of settings.
Given the efficacy of the pseudolabeling method, we are planning to publicly release the generated captions.
\end{abstract}

%%%%%%%%%%%%%%%%%%%%%%%%%%%%%%%%%%%%%%%%%%%%%%%%%%%%%%%
% Introduction
%%%%%%%%%%%%%%%%%%%%%%%%%%%%%%%%%%%%%%%%%%%%%%%%%%%%%%%
\section{Introduction}

Large language models have revolutionized natural language processing \cite{radford2019language,brown2020language_gpt3,Chowdhery2022PaLMSL} and are rapidly affecting adjacent fields such as computer vision \cite{clip,Jia2021ScalingUV,Wang2022UnifyingAT,Wang2022SimVLMSV,Yu2022CoCaCC}.
For example, using \textbf{only} weakly-supervised image-text data CLIP \cite{clip} and CoCa \cite{Yu2022CoCaCC} outperform ResNets \cite{He2016DeepRL} on ImageNet.
% The flexibility of anything-to-text \cite{Alayrac2022FlamingoAV} approach allows to have universal models that can process all text, images, and videos \cite{Lu2022UnifiedIOAU}.
Recent works also demonstrate that the flexibility of the language modeling approach allows us to apply it to any modality \cite{Alayrac2022FlamingoAV,Lu2022UnifiedIOAU}.
Nevertheless, to pre-train such models, we need enormous amounts of aligned data, which is not yet easily available for all modalities.
This holds true for most of the pre-training methods: either contrastive \cite{clip}, discriminative \cite{oscar,vilbert} or generative \cite{cho2021unifying,Alayrac2022FlamingoAV,lavender,Wang2022GITAG}.
% However, most of the pre-training methods, either contrastive \cite{}, discriminative \cite{oscar,vilbert} or generative \cite{cho2021unifying,Alayrac2022FlamingoAV,lavender,Wang2022GITAG} require enormous amounts of aligned data, which is not yet easily available for all modalities.
% Huge text collections like The Pile \cite{gao2020pile} and enormous image captioning corpora like LAION-5B \cite{schuhmann2022LAION5b} demonstrate the efficacy and scalability of web-mining,
Web-mining proved to be an invaluable source of image-caption pairs \cite{sharma2018conceptual,Changpinyo2021Conceptual1P,schuhmann2022LAION5b} due to its scalability,
but the video domain still suffers from the scarcity of aligned video-text data.

\begin{figure*}
    \centering
    \includegraphics[width=\textwidth]{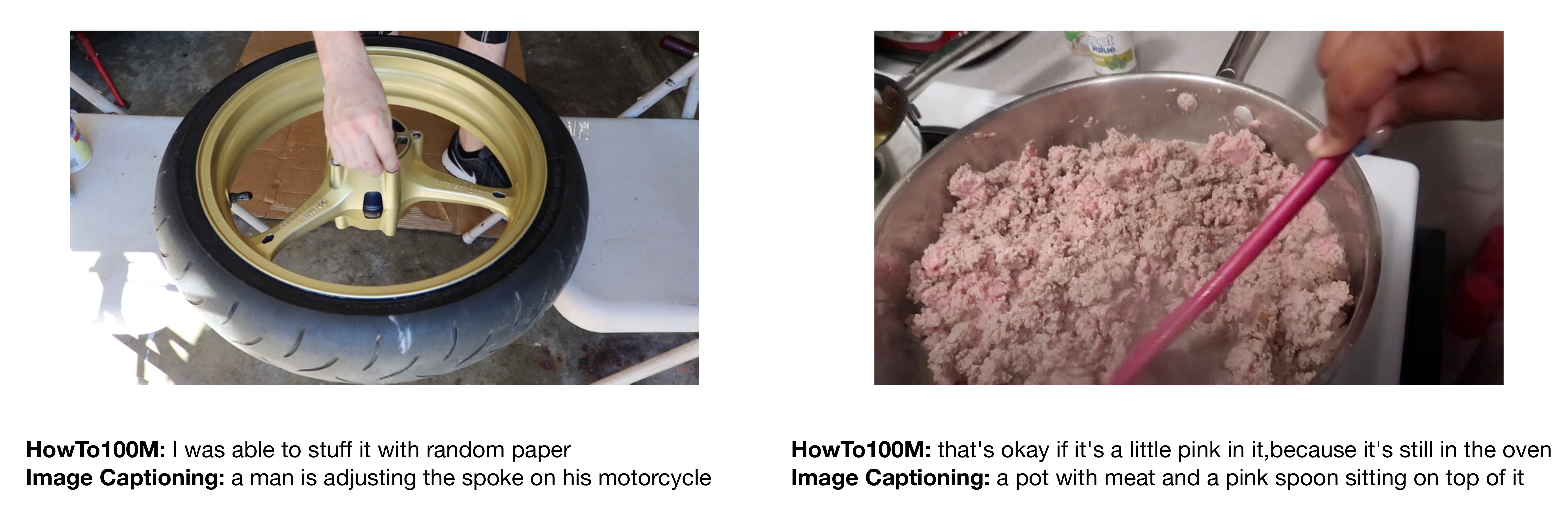}
    \caption{Image captioning models provide better descriptions than the original HowTo100M labels.}% captions often provide a poor scene description and are prone to ASR errors.}
    \label{fig:howto100m_captions}
\end{figure*}

While video data is abundant on the internet, it is hard to utilize it for pre-training.
% Especially for conditional language modeling that gains in popularity \cite{Alayrac2022FlamingoAV,lavender,Wang2022GITAG} due to its flexibility \cite{Lu2022UnifiedIOAU}.
Existing large-scale video classification datasets like Kinetics \cite{kinetics} and YouTube8M \cite{youtube_8m} consist of 500K+ video clips. Still, they only provide class labels and can not be used for generative modeling.
% that becomes more popular and successful nowadays \cite{Alayrac2022FlamingoAV,lavender,Wang2022GITAG}.
Mining videos with the alt text HTML attribute that provides a short description of the media content \cite{Bain2021frozen,Li2016TGIFAN} is a promising direction that has been immensely successful in the image captioning domain
\cite{clip,sharma2018conceptual,Changpinyo2021Conceptual1P,desai2021redcaps,schuhmann2022LAION5b}.
However, motivated by the VirTex finding \cite{Desai2021VirTexLV} that dense image annotations (captions) work better for pre-training than sparse annotations (class labels), we speculate that video alt text does not describe long videos with enough detail and is not well-suited for pre-training.
% of long videos excibit the same problem
% They do not describe the long videos with enough details.
% Processing long videos is computationally expensive and computation/training signal trade-off is especially noticeable in the video domain.
% Further, because of this, alt text mining can amplify single-frame bias in video processing~\cite{Lei2022RevealingSF}.

Nagrani et al. \cite{Nagrani2022LearningAM} propose to align existing image captions with videos by searching for video frames similar to an annotated image. For example, if an image has the caption ``pop artist performs at the festival,'' it is possible that this image is a part of a music video. Using an image as a proxy allows us to mine for aligned video-text data. Although this approach is interesting, it is limited to the videos that happen to have some of their frames labeled for image captioning. Additionally, producing such data requires an expensive pre-processing step that includes encoding multiple video frames of each video and all images, building a maximum inner-product search index, and performing the search.

Another way to get aligned text-video pairs is automatic speech recognition. Unlike alt text, it produces long, dense captions for every video. It is used in the largest video description dataset currently available -- HowTo100M \cite{miech19howto100m}.
This dataset contains 100 million instructional video clips with ASR captions.
The authors of HowTo100M specifically selected the instructional domain to better align the ASR text and the video content. For example, in this domain, an espresso making tutorial can include a caption like
\textit{``grind 18 grams of coffee beans''}, providing a weak training signal for both action recognition \textit{``grind''} and an object recognition \textit{``coffee beans''}.
However, this motivating example does not describe the usual case (Figure \ref{fig:howto100m_captions}). Using a random sample of 100 HowTo100M clips, we estimate that only 45\% of the captions refer to the video content in any ways (e.g., mention an object or an action). 13\% are intro and outro-related speech \textit{``hey guys we're back with another cooking video''} and 42\% can be best described as chit-chat \textit{``that's okay if it's a little pink in it, because it's still in the oven''}\footnote{The meat was not in the oven, this is an ASR error.}.

In this work, we propose to exploit recent advancements in image captioning \cite{Desai2021VirTexLV,li2022blip,Yu2022CoCaCC} and large-scale image-text dataset mining \cite{sharma2018conceptual,schuhmann2022LAION5b} for multimodal video pre-training. We explore how one can pre-train video captioning models without any aligned video-text data and show that pseudolabeling videos with image captioning models provides a strong baseline for building large-scale video-caption datasets.
Unlike alt text mining, a pseudolabeling approach allows for the creation of dense labels for long videos via chunking them into small clips and labeling these individually.
This approach is not limited to any particular video domain (unlike \cite{miech19howto100m}) and is computationally cheaper than the approach from \cite{Nagrani2022LearningAM}. It only requires the generation of captions for several frames, without the need for additionally encoding large image-captioning datasets and large-scale search structures.

\textbf{Our results can be summarized as follows.}
We utilize image captioning models to pseudolabel video pre-training data and show that it is possible to pre-train high-quality video models without any parallel video-text data.
Further, we demonstrate that image captioning pseudolabels work better for pre-training than the original HowTo100M ASR captions.
We investigate the importance of pre-training on both images and video and show that such a mix produces significantly better network (+4~CIDER on MSR-VTT) than pre-training on a single modality.
We introduce a new \textit{separable cross-attention} mechanism that allows to effectively attend to multidimensional data.
Finally, we describe additional unexpected findings from training large multimodal models. They include tips on adapter gate implementation and initialization and the effect of ADAM's second momentum hyperparameter on training stability.
Our methods are complementary to the existing pre-training or data mining approaches and can be used in a variety of settings.

\begin{figure*}
    \centering
    \includegraphics[width=\textwidth]{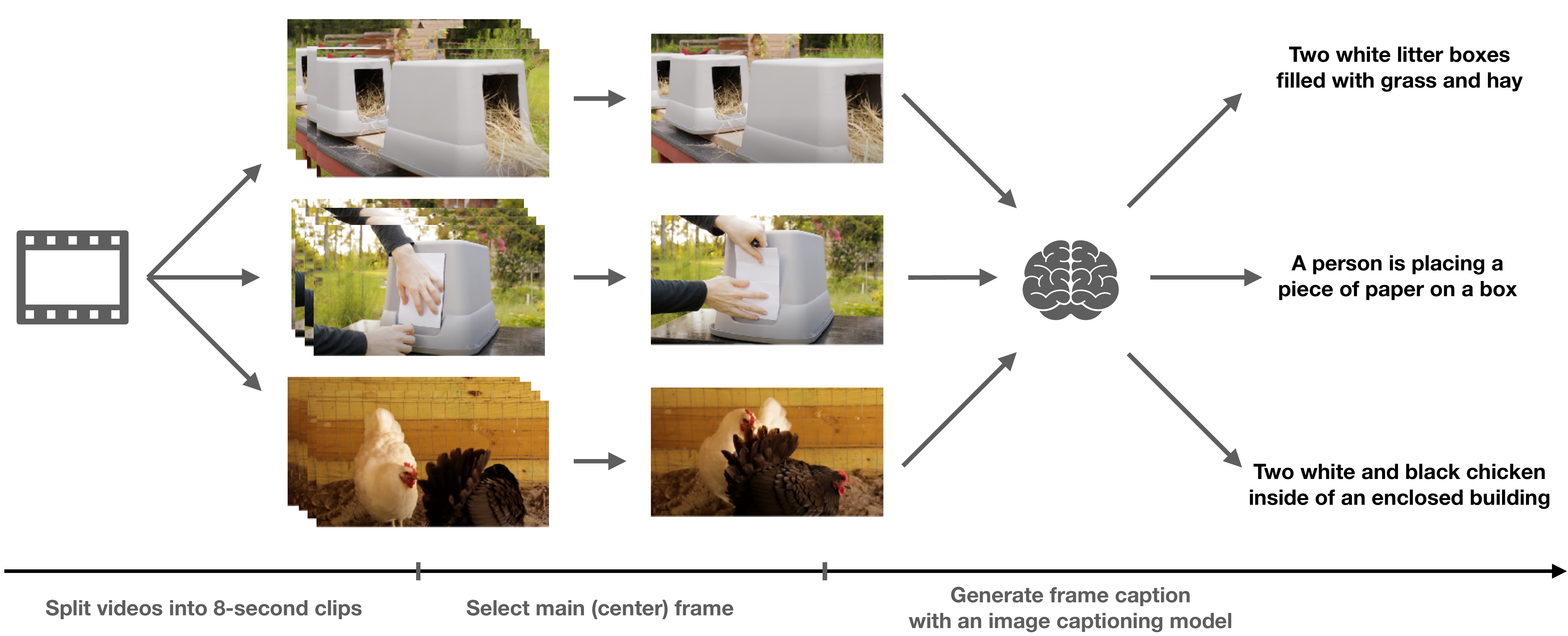}
    \caption{We use unlabeled videos and apply image captioning models to produce pre-training labels.}
    \label{fig:pseudolabeling}
\end{figure*}

%%%%%%%%%%%%%%%%%%%%%%%%%%%%%%%%%%%%%%%%%%%%%%%%%%%%%%%
% Related Work
%%%%%%%%%%%%%%%%%%%%%%%%%%%%%%%%%%%%%%%%%%%%%%%%%%%%%%%
\section{Related work}

\paragraph{Language-vision pre-training}

A pre-trained image encoder was used even in now-classical image captioning models \cite{Lei2022RevealingSF,Johnson2016DenseCapFC}, but learning a joint visual-language representation became common after the success of pre-training in NLP \cite{elmo,ulmfit,gpt,bert}.
BERT and masked language modeling (MLM) inspired a new generation of models that use self-supervised objectives to connect language and vision \cite{vilbert,lxmert,oscar,uniter,Kim2021ViLTVT,lavender}.
These methods allow learning from multimodal data using MLM-like and contrastive or optimal transport objectives.
Now, generative language modeling approaches are becoming more common in both language \cite{brown2020language_gpt3,Chowdhery2022PaLMSL} and vision \cite{Wang2022SimVLMSV,Yu2022CoCaCC,Alayrac2022FlamingoAV} thinning the lines between the multimodal NLP and computer vision fields.

% \paragraph{Video pre-training on unaligned data}
% Castro2022FitCLIPRL, \cite{Nagrani2022LearningAM},

\paragraph{Web-scale datasets}
Incredible results do not come from modeling alone. Increasing the training data amount is essential to achieve predictable improvement in performance \cite{scaling_laws}.
Internet mining is a promising approach because
unlabeled or weakly-labeled data is abundant on the Internet.
Using Wikipedia, Common Crawl, and other text sources for dataset mining was essential in the recent NLP progress \cite{bert,brown2020language_gpt3}.
CLIP \cite{clip} and ALIGN \cite{Jia2021ScalingUV} demonstrated the importance of the scale of weakly-supervised data for images.
% However, the situation in the video domain is different.

\paragraph{Large-scale video datasets}
Similar to the image domain \cite{deng2009imagenet}, until recently supervised pre-training on human-labeled datasets like Kinetics \cite{kinetics} dominated the field \cite{Zhang2020ObjectRG,Zhu2019MultiViewFA,Lin2021SwinBERTET}. However, nowadays, generative approaches are becoming more widespread and successful \cite{tang-etal-2021-decembert,lavender,Wang2022GITAG}.
These approaches require way more pre-training data than the supervised methods. Adapting image caption mining for videos is a promising direction to achieve this. Video metadata such as alt text or description of a YouTube video have been explored by Pan et al. \cite{Pan2020AutocaptionsOG} and Stroud et al. \cite{Stroud2020LearningVR}.
However, this unavoidably biases the dataset collection process towards short videos, as a single short caption cannot provide dense information about a video's visual content.
For example, Auto-captions on GIF \cite{Pan2020AutocaptionsOG} limit the number of frames in a GIF video to 50 and WTS-70M \cite{Stroud2020LearningVR} randomly select only 10 seconds of a video to download.

HowTo100M \cite{miech19howto100m} takes a different approach. Each video is automatically captioned using an ASR system. This provides diverse, dense captions, yet it creates a number of method-specific problems.
First, it restricts any video pre-training method from using both visual and audio modalities, as such models could ignore visual modality and focus on the speech alone.
Second, ASR systems introduce multiple kinds of errors specific to these systems.
Examples include mixing the speech of multiple people together, increased word error rate in noisy environments, and biasing the dataset towards people with particular accents \cite{Koenecke2020RacialDI}, propagating ASR racial biases into new datasets.
Finally, our analysis (Section \ref{sec:howto100m_analysis}) suggests that even for HowTo100M instructional videos, ASR captions \textit{often do not describe neither the scene nor the actions}.

% If the s one of the inputs as predicting video caption in this case would be reduced to learning an ASR system and potentially ignoring the visual modality completely.

\paragraph{Utilizing image captioning for the benefits of video}

Several recent state-of-the-art video understanding models \cite{Alayrac2022FlamingoAV,lavender,Wang2022GITAG} use both image captioning and video captioning datasets during pre-training.
However, the value of having both images and videos in pre-training has not yet been explicitly quantified.

Nagrani et al. \cite{Nagrani2022LearningAM} use image as a proxy between text, video, and audio. They use Conceptual Captions \cite{sharma2018conceptual,Changpinyo2021Conceptual1P} and 150M video clips and apply image search to align text to video clips and audio. Their approach yields a dataset of 6.3M video clips and 970K captions.
One can see this approach as a $k$ nearest neighbors video captioning model with $k=1$ and a similarity metric defined by the image similarity model. It lacks caption diversity, as it cannot produce new captions unseen in the training data.
In contrast to that, we propose to directly apply a high-quality image captioning model to a video frame.
This significantly reduces encoding costs as we only need to process one frame per clip and removes image-video matching costs. Unlike the KNN method, frame captioning provides more diverse video captions and allows one to apply this method to a video of any domain.
We demonstrate that this approach is simple and effective at producing large-scale video pre-training data.

% Video datasets: Howto100M, Ghadiyaram2019LargeScaleWP hashtags (IG-Kinetics-65M), Stroud2020LearningVR (WTS-70M), note that HowTo100M only has 1 million unique videos, they are just chunked into 100M videos.

% Video: Simonyan2014TwoStreamCN,Zhang2021CotrainingTW
% Video pre-training: mvgpt,lavender,git,decembert

%%%%%%%%%%%%%%%%%%%%%%%%%%%%%%%%%%%%%%%%%%%%%%%%%%%%%%%
% Methods
%%%%%%%%%%%%%%%%%%%%%%%%%%%%%%%%%%%%%%%%%%%%%%%%%%%%%%%
\section{Method}

We propose to utilize unlabeled videos and weakly-labeled image-caption pairs to construct a large-scale video captioning dataset.
In this section, we describe our method in detail. Section \ref{sec:model} describes an adapter-based video-conditioned language model and a novel separable cross-attention mechanism.
To demonstrate the method's efficacy, we pre-train our model on different datasets and evaluate them in Section \ref{sec:experiments}.

\begin{figure*}
    \centering
    \includegraphics[width=\textwidth]{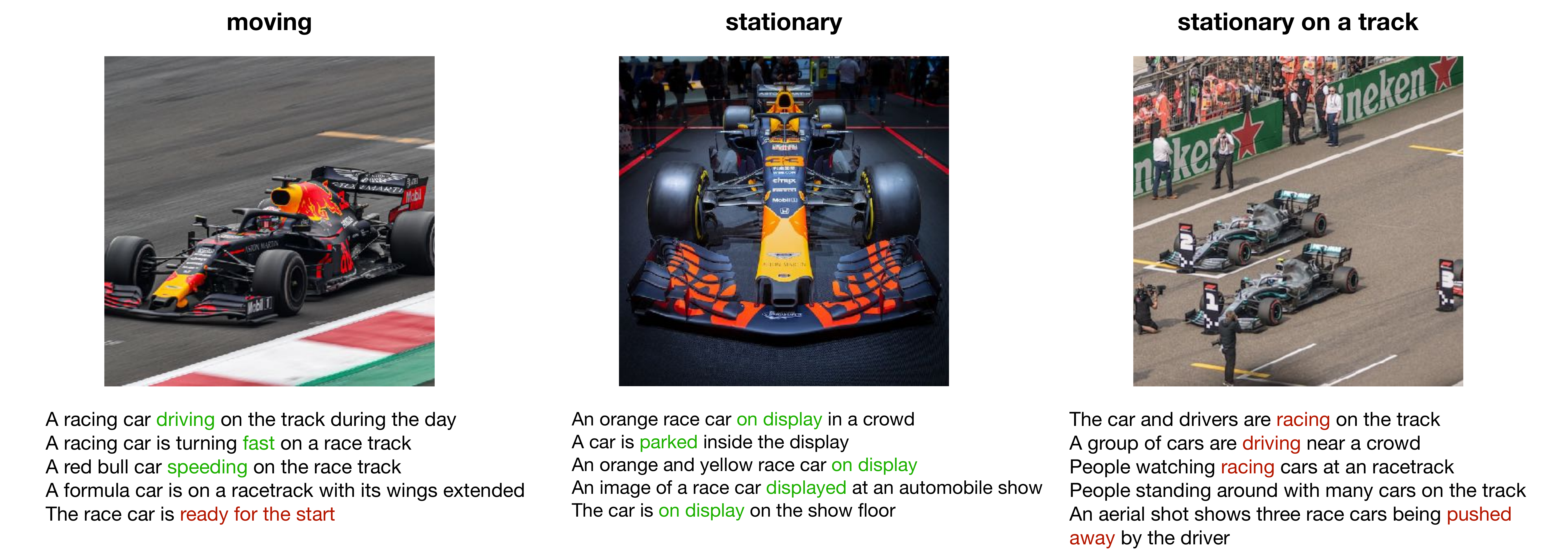}
    \caption{Image captions produced by BLIP \cite{li2022blip}. In many cases, the model can correctly recognize the action and distinguish between moving vs. stationary objects on a still image using the surrounding context.}
    \label{fig:image_captions_action}
\end{figure*}

\subsection{Pseudolabeling via image captioning}

An image captioning model can only describe a still image. It cannot \textit{explicitly} process any temporal information, such as how the objects move. However, we notice (Section \ref{sec:howto100m_analysis}) that in many cases images contain this information \textit{implicitly}.
For example, object orientation, placement, and blur allow inferring the movement.
A type of object and its relative position to other objects allow inferring the action.
In general, there is a lot of implicit information on the image that serves as a proxy allowing to describe a short clip with a single frame correctly.

% Implicit information about the movement like object orientation or implicit information about the performed action like what these objects are usually used for or the background can serve as a proxy allowing to correctly describe a short clip with a single frame.

We suggest that current image captioning models can utilize this information.
% In the left example (Figure \ref{fig:howto100m_captions}), we can see that a generated image caption correctly recognizes not just the objects, but also the action (adjusting the spoke) that is performed in a video.
For anecdotal evidence, we produce multiple captions\footnote{top-$p$ sampling with $p=0.9$} for three images with a moving race car, a standing race car, and a race car standing on a track (Figure \ref{fig:image_captions_action}). The model is able to distinguish between moving and standing based on the car's surroundings and only fails in the third case.
% While image captioning models are not perfect and still sometimes produce factually incorrect image descriptions, neural networks demonstrated high noise robustness at scale \cite{Xiao2015LearningFM,Zhang2021UnderstandingDL}.

Therefore we propose to use image captioning to produce a large video-text dataset suitable for pre-training.
Figure \ref{fig:pseudolabeling} describes our method. It consists of three steps: split videos into short clips, select the main frame and generate the frame caption.
% As a single frame can only caption video very locally
We split the videos into 8-second clips and feed the center frame into the video captioning model.
We use publicly available state-of-the-art BLIP \cite{li2022blip} model and nucleus sampling to generate captions. We do not use beam search as it tends to produce more bland and less diverse texts \cite{Holtzman2020TheCC, chan2022distribution}, which would be undesirable for pre-training.
Using a simple heuristic of selecting the center frame of a clip
% instead of ranking them via a video keyframe model
allows to minimize video decoding time by order of magnitude. This is important for high-resolution videos, where decoding can bottleneck the captioning pipeline.
For a fair comparison, we use HoTo100M videos and compare our pseudolabeling method with HowTo100M ASR captions that are commonly used to pre-train video captioning networks \cite{Li2020HeroHE,Miech2020EndtoEndLO,Seo2021LookBY,Sun2019VideoBERTAJ,Zhu2020ActBERTLG,mvgpt,tang-etal-2021-decembert}.

%%%%%%%%%%%%%%%%%%%%%%%%%%%%%%%%%%%%%%%%%%%%%%%%%%%%%%%
% Model
%%%%%%%%%%%%%%%%%%%%%%%%%%%%%%%%%%%%%%%%%%%%%%%%%%%%%%%
\section{Model}
\label{sec:model}

Our model architecture generally follows Flamingo \cite{Alayrac2022FlamingoAV}, which was chosen for the sake of simplicity and flexibility; with several modifications specific to the video modality.

We utilize a pre-trained Transformer \cite{vaswani2017attention} language model and a pre-trained TimeSformer \cite{Bertasius2021Timesformer} network.
We introduce multimodal adapters to some of the Transformer layers to condition the language generation on TimeSformer's last layer hidden states.
The weights of the Transformer and the TimeSformer are kept frozen during both pre-training and fine-tuning, we only train the parameters of the adapters.
Instead of using Perceiver \cite{Jaegle2021PerceiverGP} to resample videos we attend to a full video tensor with a novel separable cross-attention mechanism (Section \ref{sec:separable_cross_attention}). In our preliminary results we found it to be significantly faster and easier to train than Perceiver while maintaining the same language modeling performance.

We optimize conditional language modeling objective. During caption generation at any timestamp, the model can access all video frames $V$.
\begin{equation*}
\mathcal{L} = - \sum_{i = 1}^{N} \log P (w_{t} | w_{<t}, V), \text{ } V \in \mathbb{R}^{t \times h \times w}
\end{equation*}

\subsection{Multimodal adapters}

Connecting frozen models with adapters \cite{houlsby2019adapters,Alayrac2022FlamingoAV} is beneficial for several reasons.
First, adapters allow to achieve the same levels of performance as full fine-tuning of the language model \cite{houlsby2019adapters}.
Second, having less trainable parameters reduces memory requirements for optimizer states and communication volume between the GPUs in a distributed setup allowing them to scale more efficiently.
Third, the frozen visual encoder does not require backpropagating to it, which saves both memory and time. This also allows us to cache visual features during training.
Finally, using pre-trained language and vision models reduces training times. It maximizes the amount of pre-training data the model saw as a whole, which means that both vision and language models were already trained on vast amounts of unimodal data.

\begin{figure}
    \centering
    \includegraphics[width=0.5\textwidth]{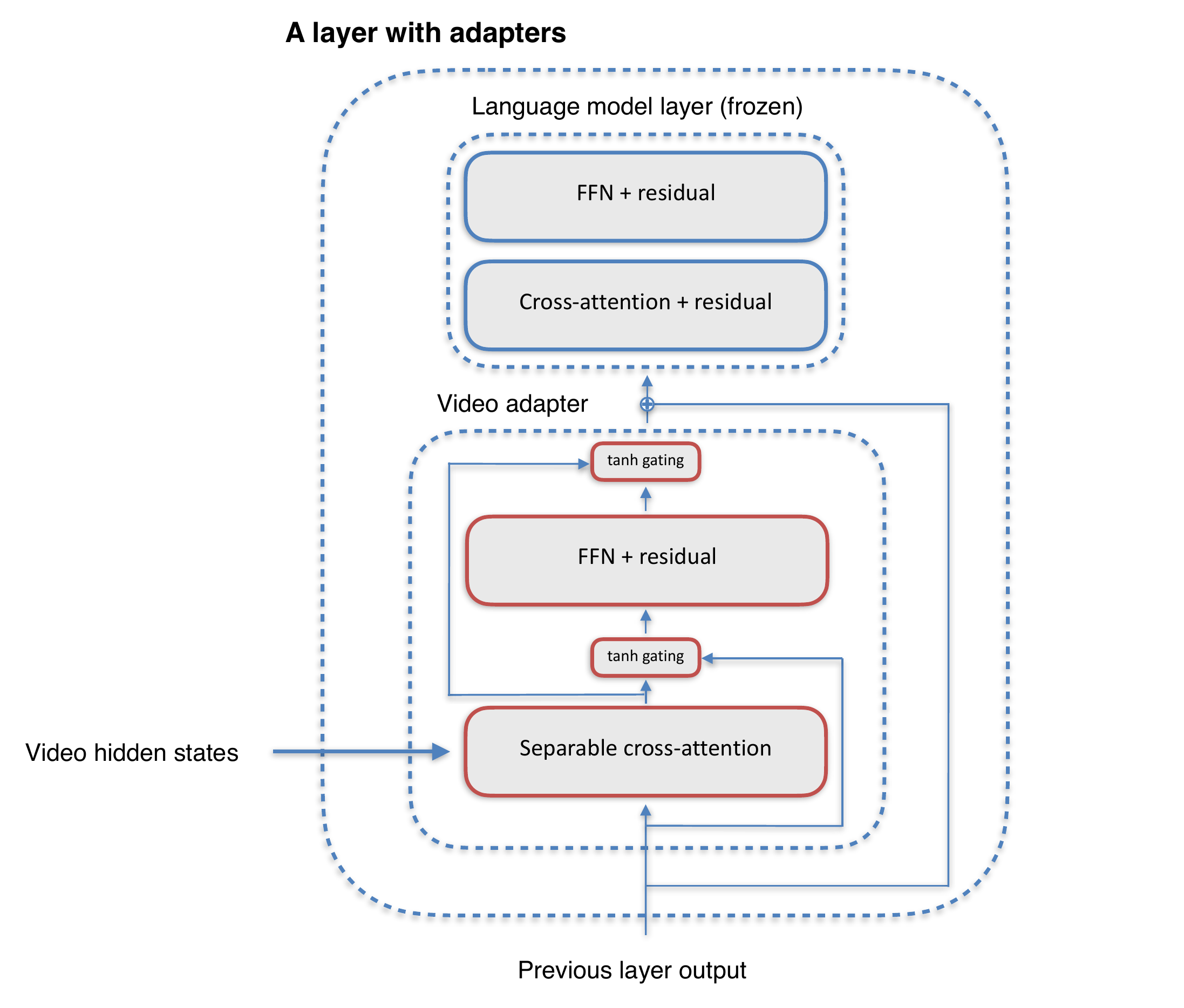}
    \caption{We insert adapters into some Transformer Layers to allow them to cross-attend to video representations using separable cross-attention.}
    \label{fig:adapter}
\end{figure}

Each adapter consists of a separable cross-attention layer and a fully-connected network similar to Flamingo \cite{Alayrac2022FlamingoAV} .
We use a pre-norm architecture and introduce $tanh$ gates to the residual connections to give the network a simple control mechanism of how much visual and text information to pass to the next layer.
Unlike Flamingo, which uses a single scalar to weight the output of a sub-layer, we find that per-dimension (vector) gates improve training stability and allow to use higher learning rates. Figure \ref{fig:adapter} summarizes the architecture.

\subsection{Separable cross-attention}
\label{sec:separable_cross_attention}

Video is a challenging modality for many reasons, but the most straightforward one is the size.
A video consists of many visual frames and naively could be represented as a tensor of shape~\texttt{[3,~t,~h,~w]}. Visual transformers \cite{Dosovitskiy2021AnII}, while being very successful at processing visual data, are extremely memory-hungry. 
% Without any architecture optimizations ViT would need to compute attention between every 16x16 patch of every timeframe of a video, which gives $6,400$ patches for a video of size 16x320x320.
The complexity of the self-attention operation is $O((t s)^2)$ where $t$ is the number of video frames, and $s$ is the number of spatial tokens.
Such computation is both costly and vastly inefficient. Multiple solutions were proposed to mitigate this problem, and one of them is separable (axial) attention \cite{Ho2019AxialAI,Bertasius2021Timesformer}.
Separable attention similarly to separable convolution \cite{Chollet2017SeparableConvolution} decomposes attention into multiple attentions of different axes. For example, time axis and space (height and width) axis. This reduces $O((t s)^2)$ to $O(t^2 s + t s^2)$ or $~40M$ operations to only $~2.6M$ in a 16-patch 16-frame video.

Nevertheless, a direct adaptation of separable self-attention to the cross-attention case is just as ineffective as full cross-attention.
Say the query size is $q$. Na\"ively cross-attending from this query to every patch of the video would give us the complexity of $O(qst)$. Straightforward application of axial attention via a reshape of the time and space tensor gives us $t$ attentions across space and $s$ attentions across time. This yields a complexity of $O(qs \cdot t + qt \cdot s) = O(2 qst)$. Such an operation is twice as expensive as a vanilla cross-attention because one has to compute multiple time and space cross-attentions.

This is why we propose to modify the separable attention mechanism for the cross-attention case.
Before computing time attention, we maxpool the video tensor over the space dimension. The same operation is performed for the space attention, but we maxpool time dimension before computing it. This gives the complexity of $O(qs + qt)$.
After that, both attention layer outputs are concatenated across the hidden dimension and downsampled using a linear layer. We additionally layer-normalize hidden states after maxpooling.
% Separable cross-attention is summarized in Algorithm \ref{alg:separable_cross_attention}.

\section{Experimental setup}
\label{sec:experiments}

\subsection{Dataset generation}

We apply our video pseudolabeling method described in Section \ref{fig:pseudolabeling} to HowTo100M videos.
All the videos are chunked into 8 seconds, and their center frame is captioned with a BLIP-Large\footnote{\texttt{BLIP/models/model\_large\_caption.pth}} model.
In order to get high-quality captions, we use an image resolution of 320x320.
Generation process took one day on 64 V100 GPUs.
The resulting dataset consists of almost 50 million 8-second clips with average caption length of 10.0 words.
To compare it to the original HowTo100M ASR captions, we manually evaluate 100 clips and find that in 65\% of the cases, pseudolabels provide better video descriptions than ASR. We also find that ASR captions relate to the video in only 45\% of the cases, while image captions correctly refer to the object and actions in the video in 88\% of the cases.

In the following sections, we compare this dataset to the original HowTo100M captions produced by an ASR system. We pre-train our model on several data variations, including the original HowTo100M dataset and LAION-5B images, and compare them on video captioning tasks.
% and \todo{question answering tasks}.

\subsection{Pre-training}

General model architecture is described in Section \ref{sec:model}. In this section we describe the particular pre-training experiments and some hyperparameter choices.

\paragraph{ASR Captions vs Image Captions}
\label{sec:howto100m_analysis}

First, we want to learn if pseudogenerated captions are better suited for pre-training than commonly used HowTo100M ASR captions.
We pre-train two models: one on 8-second clips\footnote{If needed we concatenate multiple HowTo100M clips together to form videos of roughly 8 seconds.} labeled with image captioning and one on the same clips but with the original ASR captions.
Both models are pre-trained for 4K iterations with batch size 1.2K (about 1 epoch) and then fine-tuned on MSR-VTT.

\paragraph{Pre-training on image-only data}

Given that image captioning models are trained on weakly-supervised data and then applied to videos, a natural question arises: can we pre-train a video captioning model on images only?
We treat images as one-frame videos and pre-train another model on a random subset of LAION-5B (only English captions) for the same number of steps as our video models.

\paragraph{Pre-training on a mix of data}

For this experiment, we sample examples from LAION with a probability of 0.95 and from HowTo100M with a probability of 0.05. Because preparing a video batch is usually about 20 times slower than preparing an image batch, this achieves high training throughput while having significant exposure to video data.
This model is trained for the same amount of data to make it comparable to other models.

% \begin{table}[]
%     \centering
%     \begin{tabular}{l|c}
%         % \toprule
%          & MSR-VTT \\
%         \midrule
%         FrozenCaptioner & 54.0 \\
%         -HowTo100M & 49.6 \\
%         -LAION & 49.7 \\
%         -LAION - BLIP labels & 49.0 \\
%         % \bottomrule
%     \end{tabular}
%     \caption{Video captioning results (validation set, no beam search, all models are pre-trained for 4K iterations), CIDER-D scores. ``-LAION - BLIP labels'' is pre-training on plain HowTo100M.}
%     \label{tab:video_captioning}
% \end{table}

\begin{table}[]
    \centering
    \begin{tabular}{c|c|c|c}
        % \toprule
        Image Captions & LAION-5B & ASR & MSR-VTT \\
        \midrule
            &     & $\checkmark$ & 49.0 \\
            $\checkmark$ &  &     & \underline{49.7} \\
            & $\checkmark$  &  & 49.6 \\
            $\checkmark$ & $\checkmark$ & & \textbf{54.0} \\
        % \bottomrule
    \end{tabular}
    \caption{Ablation studies. MSR-VTT validation set, no beam search, all models are pre-trained on 500K examples (videos + images if any). Pseudo-labeling is significantly more effective than original HowTo100M ASR captions. Training on only images or only videos is significantly less effective than training on both with 95\% images 5\% videos mixture.}
    \label{tab:ablation_studies}
\end{table}

% \begin{table}[]
%     \centering
%     \begin{tabular}{c|c|c|c}
%         % \toprule
%         ASR Captions & Pseudolabeling & LAION-5B & MSR-VTT CIDER \\
%         \midrule
%         $+$ &     &     & 49.0 \\
%             & $+$ &     & 49.7 \\
%             &     & $+$ & 49.6 \\
%             & $+$ & $+$ & 54.0 \\
%         \bottomrule
%     \end{tabular}
%     \caption{Video captioning results (validation set, no beam search, all models are pre-trained for 4K iterations), CIDER-D scores.}
%     \label{tab:video_captioning}
% \end{table}

\paragraph{Training efficiency tricks}

For our pre-training experiments, we use OPT-1.3B and TimeSformer pre-trained on Kinetics and HowTo100M\footnote{\texttt{TimeSformer\_divST\_96x32\_224\_HowTo100M}}. We insert six adapters to the OPT network, specifically to layers 12, 14, 16, 18, 20, and 22.
Starting from the middle of the network allows to only compute backpropagation to the last 12 layers of the network, saving on computation and following best practices of vision-language fusion \cite{lxmert,vilbert,Nagrani2021AttentionBF}.
% We train multiple models on 64 V100 GPUs for 4K iterations and a total batch size of 1.2K.
We use float16 precision for all parameters and Deepspeed Stage 2 data-parallel for distributed training.
To maximize GPU utilization, we use roughly eight times larger batch size for images than for videos. This accomplishes two things: maximizing the batch size, thus increasing throughput and improving stochastic gradient estimate, and evening out the time for a forward-backward pass across the GPUs, minimizing wait times before synchronization.

Using a distributed training setup simplifies data mixing and batching.
Every batch contains either images or videos, but different GPUs sample if they need to process images or videos independently. This means that the stochastic gradient estimate across all GPUs includes both images and videos. At the same time, each particular GPU processes the same kind of data without needing padding or complex batching rules.
We also test fully-synchronized modality selection when all GPUs process either videos or images simultaneously, but we find fully-synchronized training very unstable compared to mixed training.

% \begin{table}[]
%     \centering
%     \begin{tabular}{l|c|c}
%         % \toprule
%         & MSVD & MSR-VTT \\
%         \midrule
%         O2NA \cite{Liu2021O2NAAO} & 96.4 & 51.1 \\
%         DECEMBERT \cite{tang-etal-2021-decembert} & - & 52.3 \\
%         MV-GPT \cite{mvgpt} & - & 60.0\footnotemark \\
%         LAVENDER \cite{lavender} & 150.7 & 60.1 \\
%         GIT \cite{Wang2022GITAG} & 180.2 & 73.9 \\
%         \midrule
%         FrozenCaptioner & 128.8 & 54.6 \\
%         % \bottomrule
%     \end{tabular}
%     \caption{Comparison with the state of the art models.}
%     \label{tab:video_captioning}
% \end{table}

\begin{table*}[]
    \centering
    \begin{tabular}{l|c|c|c|c}
        % \toprule
        & Pre-training data & Input features & MSVD & MSR-VTT \\
        \midrule
        O2NA \cite{Liu2021O2NAAO} & - & video frames & 96.4 & 51.1 \\
        DECEMBERT \cite{tang-etal-2021-decembert} & HowTo100M & video frames, ASR, image captions & - & 52.3 \\
        MV-GPT \cite{mvgpt} & HowTo100M & video frames, ASR & - & 60.0 \\
        LAVENDER \cite{lavender} & LAVENDER mixture & video frames & 150.7 & 60.1 \\
        GIT \cite{Wang2022GITAG} & GIT mixture & video frames & 180.2 & 73.9 \\
        \midrule
        FrozenCaptioner & HowTo100Mblip + LAION & video frames & 128.8 & 54.6 \\
        % \bottomrule
    \end{tabular}
    \caption{Comparison with the state of the art models. HTM stands for HowTo100M. LAVENDER mixture is Vid2.5M \cite{Bain2021frozen} + CC3M \cite{sharma2018conceptual} + CC12M \cite{Changpinyo2021Conceptual1P} + COCO \cite{Lin2014MicrosoftCC} + Visual Genome \cite{Krishna2016VisualGC} + SBU Captions \cite{sbu_captions} + 12M crawled video text pairs. GIT mixture is similar to LAVENDER, but also includes ALT200M \cite{Hu2022ScalingUV}.}
    \label{tab:video_captioning}
\end{table*}

\section{Results}
\label{sec:results}
We fine-tune pre-trained models on MSR-VTT and MSVD video captioning datasets. Comparison of different pre-training datasets is presented in Table \ref{tab:ablation_studies}.
% We fine-tune adapters of our models on MSVD and MSR-VTT datasets until convergence on validaiton set.
% We use crop size 320px and 16 frames sampled uniformly from each video. Then we use validation set to search for optimal beam size from 1 to 10 with length penalty 0.6.
Using HowTo100M ASR captions produces the worst results of all, demonstrating the low quality of video-text alignment that ASR provides.
On the other hand, our pseudolabeled dataset performs on par with a similar amount of image captioning data.
Combining videos and images outperforms the rest by more than 4 CIDER-D points, showing that having both modalities in the data is the most effective.

To push the results further, we train our model for 40K steps on the mixture of videos and texts.
Our results compared to state of the art are presented in Table \ref{tab:video_captioning}.
Our model underperforms the current state of the art that we attribute to the frozen visual network.
Unlike GIT \cite{Wang2022GITAG}, we do not modify visual representations inside a transformer but only attend to them.
This, together with separable cross-attention (Section \ref{sec:separable_cross_attention}), allows our model to scale linearly with the number of video frames and attend to significantly longer videos than GIT, which scales quadratically.

\subsection*{Additional findings}

\paragraph{Vector gates improve training stabiliy}

We observed that using scalar gates similar to Flamingo \cite{Alayrac2022FlamingoAV} causes loss divergences at high learning rates (greater than $10^{-3}$). One simple and effective way of mitigating this problem was using vector (per-dimension) gates that allowed us to use a very high learning rate $7 \cdot 10^{-3}$.
We hypothesize that per-dimension gates serve as a kind of a normalization layer \cite{Ioffe2015BatchNA,Ba2016LayerN}, and they can cut off some large value dimensions in the adapter outputs.

\paragraph{Effect of Adam second momentum}

Starting with GPT-3 \cite{brown2020language_gpt3} several large models \cite{zhang2022opt,Smith2022UsingDA} used Adam's \cite{Kingma2015AdamAM} $\beta_2 = 0.95$ which is significantly smaller than the default $\beta_2 = 0.999$.
In our experiments we found that a small $\beta_2$ value stabilizes the training with minimal effect on convergence speed. However, several of our experiments suggest that small values of $\beta_2$ can negatively impact generalization and fine-tuning capabilities. We found that models trained with $\beta_2=0.95$ for 10K+ steps can significantly underperform models trained with $\beta_2=0.999$ trained for 4K steps on downstream tasks.
% This difference in downstream performance 
This happens even though the pre-training loss of 320px models is lower, suggesting that $\beta_2=0.95$ hurts generalization capability.

\paragraph{Unreasonable effectiveness of tanh(1) initialization}

Flamingo initializes gates at $tanh(0) = 0$. This achieves two things: first, it maximizes the gradient through the $tanh$ nonlinearity allowing to learn optimal gate values faster. Second, it allows each layer to smoothly learn how much visual information it should contribute to the language model.
However, adapter values change very slowly during training, requiring many thousands of iterations to converge. This is usually way past the point of overfitting and losing generalization capabilities on relatively small downstream datasets like MSR-VTT.

\begin{figure}
    \centering
    \includegraphics[width=0.5\textwidth]{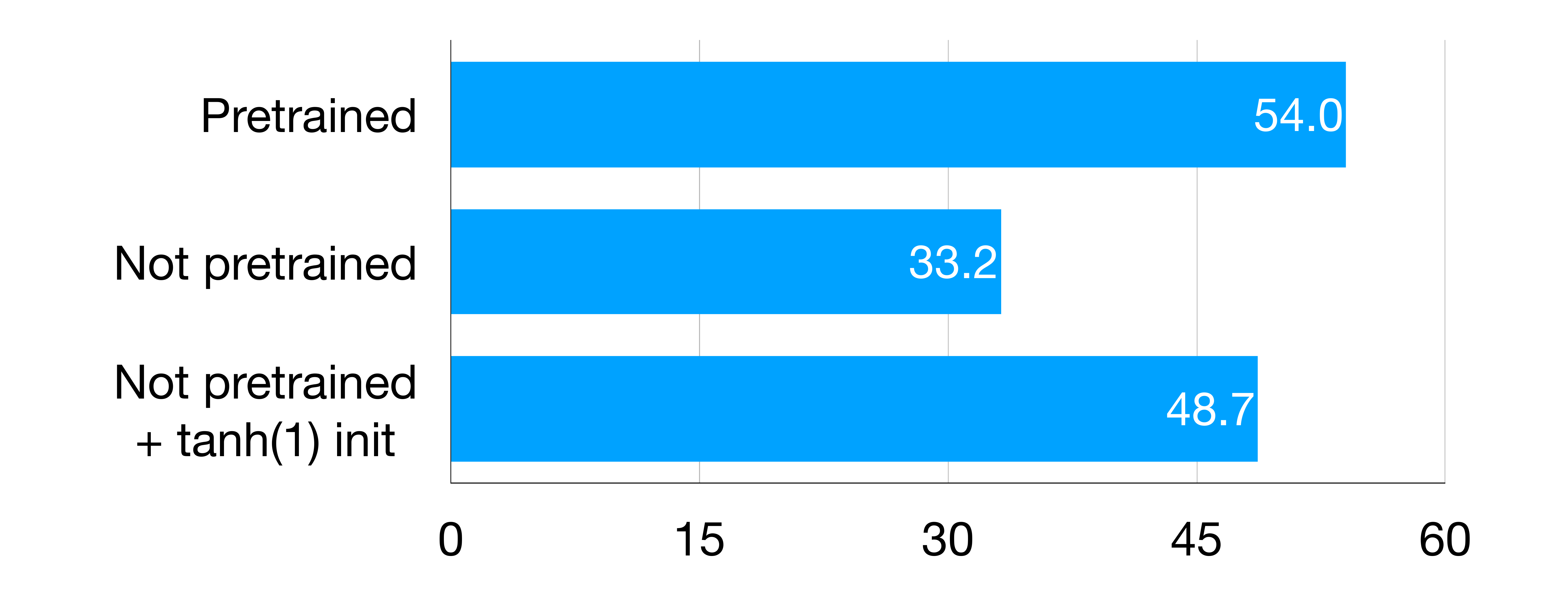}
    \caption{Initialization of visual gates closer to 1 boost non-pretrained network performance. MSR-VTT CIDEr-D scores after fine-tuning.}
    \label{fig:tanh_1_init}
\end{figure}

For a more fair comparison between pre-trained and non-pre-trained networks, we evaluate non-pre-trained in two scenarios. The adapter gates of the first network are initialized at $tanh(0) = 0$ and for the second they are initialized at $tanh(1) \approx 0.8$.
% A model with adapters initialized from scratch with $tanh(0)$ init achieves only 33.2 CIDER-D points, but a model initialized at $tanh(1)$ 
Initializing $tanh$ values closer to $1$ skyrockets a non-pre-trained models performance and reduces the pre-trained model's gap from $20.8$ to $5.3$ CIDER-D points (Figure \ref{fig:tanh_1_init}).
Diving deeper into networks fine-tuned on MSR-VTT shows almost no change with less than $0.05$ absolute change in gate values during fine-tuning for either model explaining the drastic difference in performance.

% \begin{table}[]
%     \centering
%     \begin{tabular}{l|c}
%         & MSR-VTT \\
%         \midrule
%         FrozenCaptioner & 54.0 \\
%         -pretraining & 33.2 \\
%         -pretraining + tanh(1) init & 48.7\\
%     \end{tabular}
%     \caption{Caption}
%     \label{tab:tanh_1_init}
% \end{table}

% \paragraph{Pre-training hyperparameters matter a lot} (a.k.a. gray vs green network)

\paragraph{Effect of crop size during pre-training and fine-tuning}

We pre-train two networks: with 224x224 crop and with 320x320 crop. Both of them are trained with the same batch size and the number of update steps\footnote{We compensate for increased memory using gradient accumulation}.
After pre-training we observe that the 320px crop network has lower training loss, but underperforms the 224px network on MSR-VTT consistently throughout the training.
% with the CIDER-D gap from \todo{NN} (\todo{MM} pre-trainiing steps) to \todo{NN} points (\todo{MM} pre-training steps).

An intuitive explanation could be that the TimeSformer vision encoder was pre-trained on 224x224 videos, and using a larger crop introduces a distribution shift. However, after fine-tuning, we see a different picture. Using larger video sizes during fine-tuning consistently improves captioning quality, plateauing at 320-380 pixels.
% \todo{(Figure \ref{fig:resolution})}.
Using a smaller resolution of 224x224 during pre-training, thus, seems like a great way to improve both training speed and quality, but the reason why it is so effective requires further investigation.

\section{Conclusion}

In this paper, we show that it is possible to pre-train high-quality video models without any parallel video-text data.
To do this, we employ a simple but effective video pseudolabeling technique: captioning individual frames with high-quality image description models.
% We demonstrate that current image captioning models can use image background to infer the action performed on the image providing the video dataset not just with object labels but with more dynamic information.
We demonstrate that current image captioning models
% can use image background to infer the action performed on the image providing the video dataset not just with object labels but with more dynamic information.
can provide useful video captions that allow the network to learn both static (objects) and dynamic (actions) information about the video.
% We develop a simple pre-training method and an intuitive efficient cross-attention mechanism.
To evaluate our pseudolabeling method, we pre-train several adapter-based captioning models and show that image captions provide better training signal than commonly used ASR captions. We additionally demonstrate the importance of using both images and videos in pre-training.
Finally, we develop a new cross-attetion method that allows to effectively and efficiently attend to dense video representations.

\section{Limitations}

Using unaligned video data is a promising path toward high-quality video models. However, image captioning models are not perfect. They make factual mistakes and exhibit societal biases such as race, gender, cultural and more from the training data.
Using them for large-scale dataset creation can amplify these biases and requires mitigation techniques.
Image captioning models also suffer from hallucinations. For example, mentioning objects that are not on a picture.
While a high-resolution frame that we use for captioning contains a lot of information that can be inferred, aligned video-text data mining is still an open question. A potential solution could be a combination of web-mining techniques that work well for short videos and dense pseudolabeling techniques for longer videos.

We also would like to highlight some of the domains where our methods can perform poorly.
For example, video description for hearing-impaired people. This task requires full video understanding including visual and audio modalities. Image captioning pseudolabels do not utilize audio modality and cannot provide a training signal to describe what people say or what sounds the environment makes and should be augmented with aligned (or pseudoaligned) audio-text data as well.

{\small
\bibliographystyle{ieee_fullname}
\bibliography{references}
}

%%%%%%%%%%%%%%%%%%%%%%%%%%%%%%%%%%%%%%%%%%%%%%%%%%%%%%%%%%%%

\end{document}